\documentclass{article}

\PassOptionsToPackage{numbers}{natbib}

\usepackage[final]{nips_2018}

\usepackage[utf8]{inputenc} 
\usepackage[T1]{fontenc}    
\usepackage{hyperref}       
\usepackage{url}            
\usepackage{booktabs}       
\usepackage{amsfonts}       
\usepackage{nicefrac}       
\usepackage{microtype}      
\usepackage{graphicx}
\usepackage{amsmath}
\DeclareMathOperator*{\argmax}{arg\,max}

\usepackage{caption}
\usepackage{subcaption}
\usepackage{multirow}
\usepackage{makecell}

\title{Explore-Exploit: A Framework for Interactive and Online Learning}

\author{
  Honglei Liu \\
  \And Anuj Kumar \\
  \And Wenhai Yang \\
  \And Benoit Dumoulin \\
  \AND
  Facebook Conversational AI \\
  1 Hacker Way, Menlo Park, CA 94025 USA \\
  \texttt{\{honglei, anujk, wenhai, benoitfb\}}@fb.com
}

\begin{document}

\maketitle

\begin{abstract}
Interactive user interfaces need to continuously evolve based on the interactions that a user has (or does not have) with the system. This may require constant exploration of various options that the system may have for the user and obtaining signals of user preferences on those. However, such an exploration, especially when the set of available options itself can change frequently, can lead to sub-optimal user experiences. We present Explore-Exploit: a framework designed to collect and utilize user feedback in an interactive and online setting that minimizes regressions in end-user experience. This framework provides a suite of online learning operators for various tasks such as personalization ranking, candidate selection and active learning. We demonstrate how to integrate this framework with run-time services to leverage online and interactive machine learning out-of-the-box. We also present results demonstrating the efficiencies that can be achieved using the Explore-Exploit framework.

\end{abstract}

\section{Introduction}
Continuously adapting to user's preferences is important for the success of many modern systems. For example, in recommender systems, the ultimate goal is to provide recommendations that users find useful. When a user interacts with (or ignores) a piece of content, it provides a valuable signal to improve the relevancy of the system over time. However, many modern systems have to deal with problems such as bootstrapping for new users, quickly adapting for new content that's constantly added, and ever-changing preferences of their users. 

"Exploration" and "Exploitation" are popular techniques in Machine Learning community to address these problems, but the challenges are in quickly incorporating these techniques in a production system \cite{wang2014exploration, mcinerney2018explore}. In other words, simple utilization of user feedback data (known popularly as "exploitation") might not be enough. It's equally important to proactively collect signals that would otherwise be hidden to the system with exploration. For example, showing new type of content that a user has not seen before can give additional information on a user's preference, which might be otherwise hard to obtain. Yet, we may want to balance such new unknown content, with known relevant content to maintain an overall relevancy and usability for the system. This forms the foundation of interactive and online learning, by adding meaningful user feedback in the loop. Studies have shown that with proper balance between exploration and exploitation, the performance of a recommendation system can be substantially improved \cite{wang2014exploration, mcinerney2018explore, radlinski2008learning, li2010contextual}. 

To this end, exploration in a production system must be conducted with extreme care as it could lead the system to show sub-optimal choices, thereby causing regression of user experience, i.e. regret. So, when choosing the items to be explored, we should prioritize those having a better chance of generating user feedback, as well as, those that can give us a meaningful signal that we don't already have. Therefore, when designing an interactive and online learning framework, we should consider the following requirements: (1) the framework should be able to balance exploration vs. exploitation such that we can minimize regret while getting the most information, (2) the exploration and exploitation behaviours should be highly configurable as different applications may have different needs and requirements, and (3) the framework should also be both easily attachable and detachable to any run-time service depending on product needs.

In this paper, we present Explore-Exploit: a framework that makes it easy to do online and interactive learning in production. It's designed to collect and utilize user feedback in an interactive and online fashion to minimize regret. Explore-Exploit does so by employing a suite of built-in online learning operators. It's also highly configurable to allow customizing \textit{what} targets to explore, \textit{when} to trigger the exploration,  \textit{who} will receive the exploration, \textit{which} operators to use for the exploration and \textit{where} to fetch user feedback signals for online learning. In addition, it also has a subscription based design that allows any run-time service to easily integrate. Currently, it supports a variety of tasks including ranking, candidate selection and active learning.

Specifically, Explore-Exploit has been used for the following applications:
\begin{itemize}
    \item Simple Exploration. For example, explore the ranking of suggestions to gather unbiased training data for training a ranking model
    \item Online Learning (with real-time feedback). For example, picking the winner among multiple product options or automatically fine-tuning machine learning models by choosing the best hyper-parameters.
    \item Active Learning. Explore models in an uncertainty interval and use user feedback data to improve reach and precision.
    \item Reinforcement Learning (RL). Deploy and iterate on reinforcement learning models with proper exploration.
\end{itemize}

\section{System Design}

In this section, we introduce the architecture of Explore-Exploit and motivations behind the design decisions. We also detail the operators we have built in the framework along with examples of how this framework works in practice.

\subsection{Architecture}

\begin{figure}[h]
  \centering
  \includegraphics[width=0.9\textwidth]{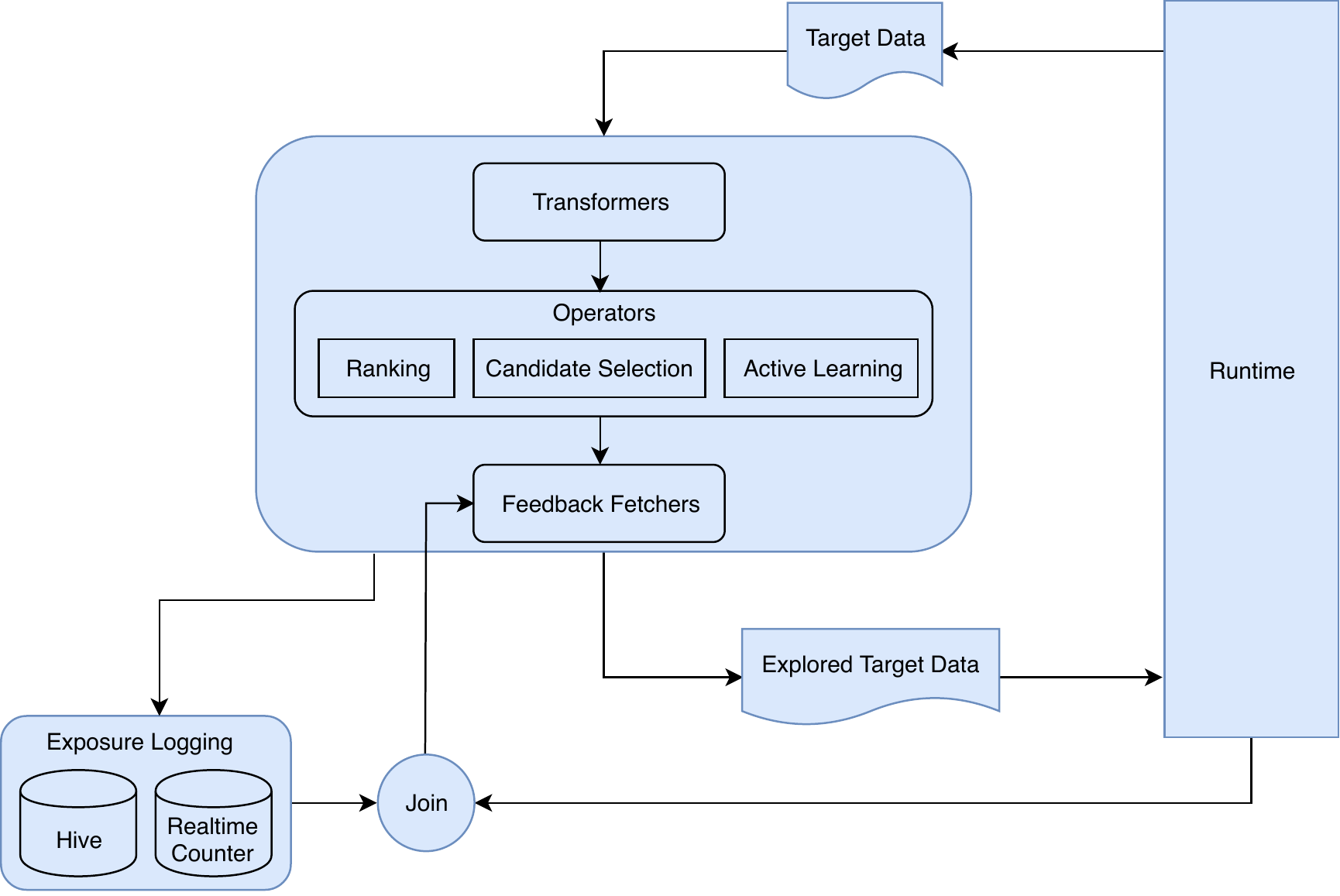}
  \caption{Architecture of the Explore-Exploit framework}
  \label{fig:design}
\end{figure}

We briefly describe the architecture of the Explore-Exploit framework in Figure \ref{fig:design}. To use the framework for any run-time service, we first add an \textit{Exploration Target} in the framework that defines \textit{Transformers}, \textit{Operators}, \textit{Feedback Fetchers} and some meta data. At any point during run-time, we can subscribe to the \textit{Exploration Target} and send \textit{Target Data} to Explore-Exploit for exploration. Then the framework will take care of logging, run online learning or interactive learning algorithms and eventually return the \textit{Explored Target Data}.

\textbf{Subscription based Service}. One consideration when building the framework was that we want to make it generalizable to any system and easy to integrate (or remove). So we designed Explore-Exploit as a subscription based service. We can send data to Explore-Exploit at any point during run-time, but the data will only be explored if we subscribe to a target defined in the framework. Otherwise, the data will be returned as it is. This allows us to easily turn on or off the interaction with Explore-Exploit as needed.

\textbf{Target Data} and \textbf{Explored Target Data}. Target data contains the data that we want to explore, such as the recommendation candidate we want to personalize and the hyper-parameter we want to automatically learn for a machine learning model. The explored target data are what will be returned by Explore-Exploit. They could have been altered from the original target data for online and interactive learning. 

\textbf{Exploration Target}. An exploration target in Explore-Exploit defines all the information needed to run the exploration, including the type of task (e.g. candidate selection or active learning), the operators we want to use, what user feedback signals we want to use for online learning, how to control the traffic, etc. One exploration target can be shared by a cluster of data sets that could be treated similarly in Explore-Exploit. For example, the hyper-parameters from multiple models can share the same target.

\textbf{Transformer}. For each exploration target, we can define a set of transformers that will be chained together and applied to the target data. Each transformer corresponds to a certain task we want to perform. For instance, learning hyper-parameters for different languages could be different transformers.

\textbf{Operator}. Each operator corresponds to an algorithm the transformer can use. Currently, there are multiple operators in Explore-Exploit divided into three types: ranking, candidate selection and active learning.

\textbf{Feedback Fetcher}. A feedback fetcher tells Explore-Exploit where to fetch user feedback. It's customizable such that we can easily choose what metric we want to optimize for through online and interactive learning, e.g., click through rate, daily active users, etc. It even gives us the flexibility to decide whether we want to do the optimization at user level or global level depending on whether the feedback is user specific.

\textbf{Exposure Logging}. All the data that are explored will be automatically logged both in Hive for offline processing and in a real-time counter service for online consumption. To get real-time feedback, the online counters can be joined with events (e.g. display and click) from the run-time service and provided to the feedback fetcher.

\subsection{Operators}

Within Explore-Exploit, we have already built 14 operators serving different purposes such as candidate selection, active learning and ranking. It's also straightforward to register and implement new operators in this framework. In this section, we describe the available operators and their functionality in Table \ref{tab:op}. 

As part of the design, for the operators with the same type, they are compatible and can be safely chained together or swapped. For example, if we want to deploy a reinforcement learning model to choose actions (e.g. showing a notification v.s. not showing), we can use the \textit{RLActionSelection} operator which will choose actions based on the predicted reward given by an RL model. However, in order to make RL models perform well in practice, we often need to do multiple rounds of interactive learning with RL policy exploration. To this end, we can easily chain a \textit{SoftmaxSelection} operator after \textit{RLActionSelection}. It will first normalize the predicted rewards with a softmax function and then sample an action from the normalized probability distribution, thus exploring the RL policy. Once the RL model reaches to production-ready quality, we can safely remove the \textit{SoftmaxSelection} operator.

\begin{center}
\captionof{table}{The list of built-in operators in Explore-Exploit} \label{tab:op} 
\begin{tabular}{ c|c|c } 
\hline
Exploration Types & Operators & Notes \\
\hline
\multirow{6}{*}{Candidate Selection} & EpsilonGreedySelection & \makecell{An operator that \\ implements the $\epsilon$-Greedy algorithm\cite{vermorel2005multi}}   \\ 
\cline{2-3}
& UCB1Enhanced & \makecell{An operator based on UCB1 algorithm \cite{kuleshov2014algorithms} \\ tailored to our needs} \\ 
\cline{2-3}
& ThompsonSampling & \makecell{An operator based on \\ Thompson Sampling algorithm\cite{agrawal2012analysis}} \\ 
\cline{2-3}
& RLActionSelection & \makecell{An operator to choose actions \\ based on reinforcement learning models \\ with flexible configurations} \\ 
\cline{2-3}
& SoftmaxSelection & \makecell{An operator to choose actions \\ based on softmax exploration strategy\cite{vermorel2005multi}} \\ 
\cline{2-3}
& BinarySearchSelection & \makecell{An operator that performs binary search \\ to find the candidate with a reward \\ closest to a target reward} \\ 
\cline{2-3}
& UniformSelection & \makecell{An operator that \\ randomly explores candidates} \\ 
\hline
\multirow{6}{*}{Active Learning} & SampleWithInterval & \makecell{Explore targets that\\ have scores in the defined intervals} \\ 
\cline{2-3}
& SampleWithIntervalDecay & \makecell{Explore targets with probabilities \\ defined by an interval decay function} \\ 
\cline{2-3}
& SampleWithEntropy & \makecell{Explore targets based on entropy} \\ 
\cline{2-3}
& SampleWithEntropyMultiClass & \makecell{Explore targets based on entropy\\ for multiclass models} \\ 
\cline{2-3}
& StratifiedSampling & \makecell{Explore targets based on \\ the distribution of predictions \\ in different intervals} \\ 
\cline{2-3}
& SampleWithSemanticSimiliarity & \makecell{Explore targets based on \\ their semantic similarities to \\ the given data points} \\ 
\hline
\multirow{1}{*}{Ranking} & ShuffleRanking & \makecell{An operator that \\ randomly explores ranking} \\ 
\hline
\end{tabular}
\end{center}

Due to space limit, we won't be able to get into details for all the operators. But here, we want to zoom into one operator, \textit{UCB1Enhanced}, to illustrate how we modified the UCB1 algorithm \cite{kuleshov2014algorithms} to fit our needs. Originally, UCB1 will select the candidate that achieves the highest upper confidence bound (UCB for short). That is, UCB1 will favor those candidates that either have high reward or high uncertainty. However, in practice, we do not always want to maximize the reward. For example, if the reward is click through rate (CTR), we may want to maintain CTR at a certain level to balance reach. Meanwhile, we also want to control how aggressive we explore those candidates with high uncertainty. In these cases, the original UCB1 algorithm won't satisfy our needs. So, we have implemented \textit{UCB1Enhanced} operator in Explore-Exploit.

Assuming we have multiple candidates (e.g. a set of thresholds) and we want to automatically choose one such that it can give us a target reward (e.g. a target CTR). Let $T$ represent the total number of rounds of interactive learning we have performed so far. Eventually, we want to select a candidate $a$ that can give us the target reward $target$. Given our current round $t$, \textit{UCB1Enhanced} will select candidates based on the following objective function

\begin{equation}
\label{eq:ucb}
    Objective= \argmax_a
    \begin{cases}
        w \times u_t(a) - \frac{1}{\delta} \left| r_t(a) - target \right|,& \text{if } r_t(a) \geq target\\
        w \times u_t(a) - \delta \left| r_t(a) - target \right|,              & \text{otherwise}
    \end{cases}
\end{equation}
where $w$ is a weight that controls the aggressiveness of doing exploration, $u_t(a)=\sqrt{\frac{2\log T}{n_t(a)}}$, $n_t(a)$ is the number of explorations for candidate $a$, $r_t(a)$ is the reward for candidate $a$ and $\delta$ is the penalty factor for small rewards compared to $target$.

\subsection{Examples}

\begin{figure}[h]
  \centering
  \includegraphics[width=0.9\textwidth]{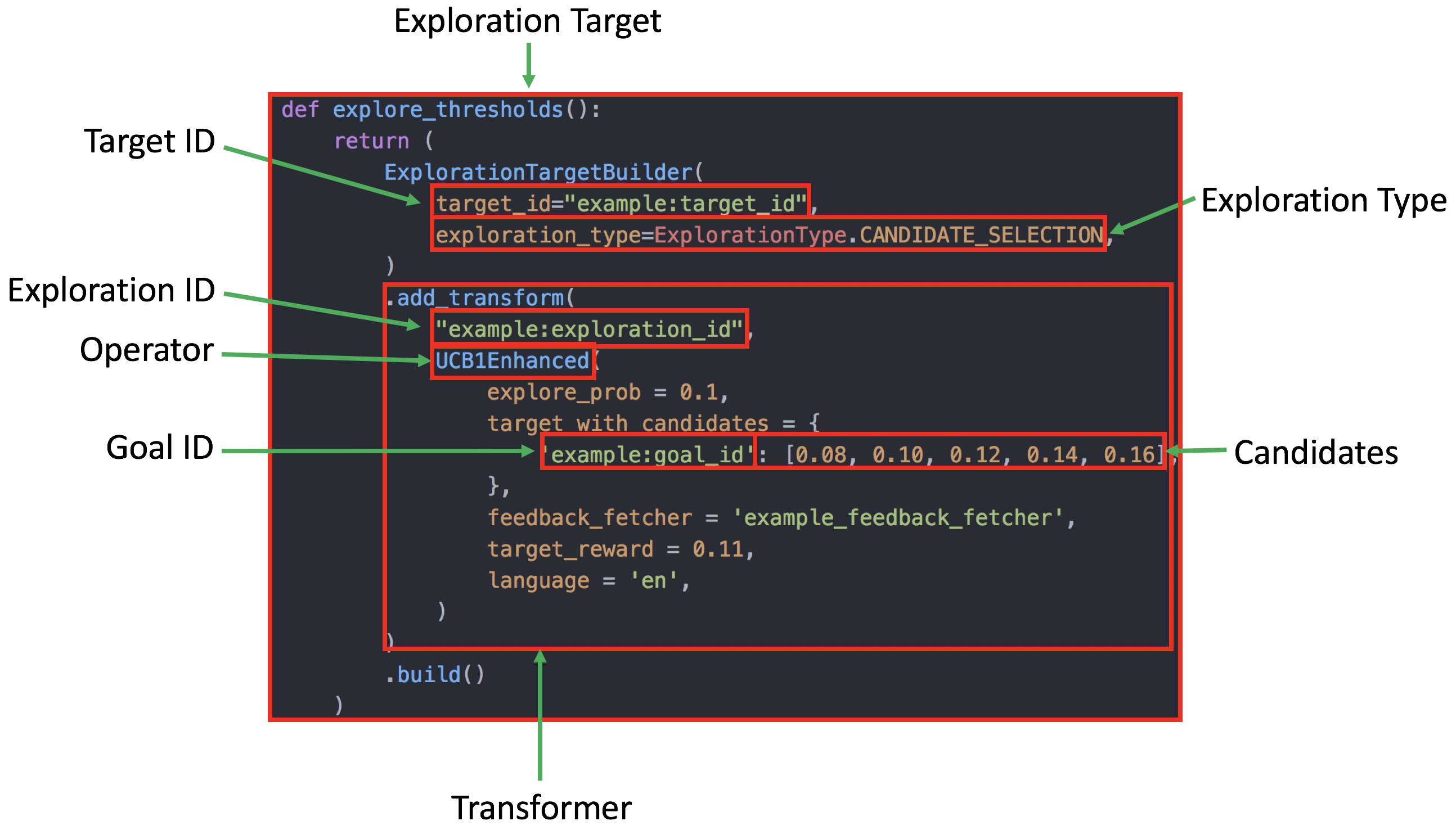}
  \caption{An example of exploration targets defined in Explore-Exploit}
  \label{fig:config}
\end{figure}

To illustrate how Explore-Exploit can be used in practice, we show an example of the exploration target in Figure \ref{fig:config}. In this example, once the target has been subscribed, the Explore-Exploit framework will explore the thresholds of the model with id \textit{example:goal\_id} with a probability of \textit{0.1} and only do the exploration if the language of the request is \textit{English}. The approach is to use the \textit{UCB1Enhanced} operator to choose the candidate that can give us a target reward of \textit{0.11} according to the feedback provided by \textit{example\_feedback\_fetcher}.

\section{Experiments}

To experiment with Explore-Exploit, we did several public tests on learning thresholds of a personalization model with the online learning operators implemented in the framework. Originally, choosing thresholds is a lot of pain since it requires an engineer to constantly try different values and closely monitor the metrics. As we scale to more and more models, the whole process may take weeks to finish. With Explore-Exploit, we make all of these easy. After a few days, the online learning operator will automatically converge to the candidate we want. It's also super configurable so we can add and delete candidates at anytime. 

\begin{figure}[h]
     \centering
     \begin{subfigure}[b]{0.5\textwidth}
         \centering
         \includegraphics[width=\textwidth]{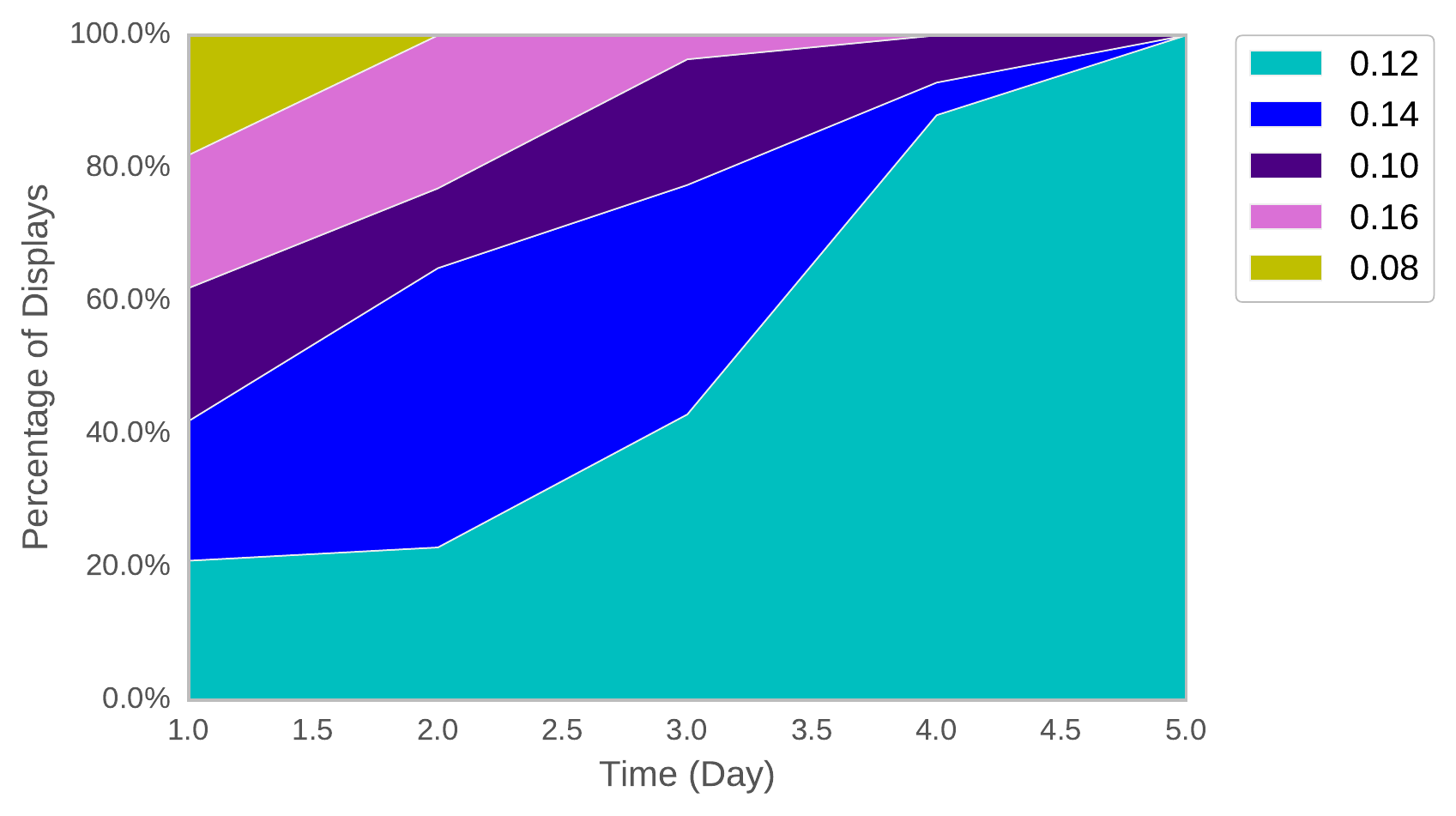}
         \caption{Percentage of displays of different candidates over time}
         \label{fig:ctr_display}
     \end{subfigure}
     \hfill
     \begin{subfigure}[b]{0.4\textwidth}
         \centering
         \includegraphics[width=\textwidth]{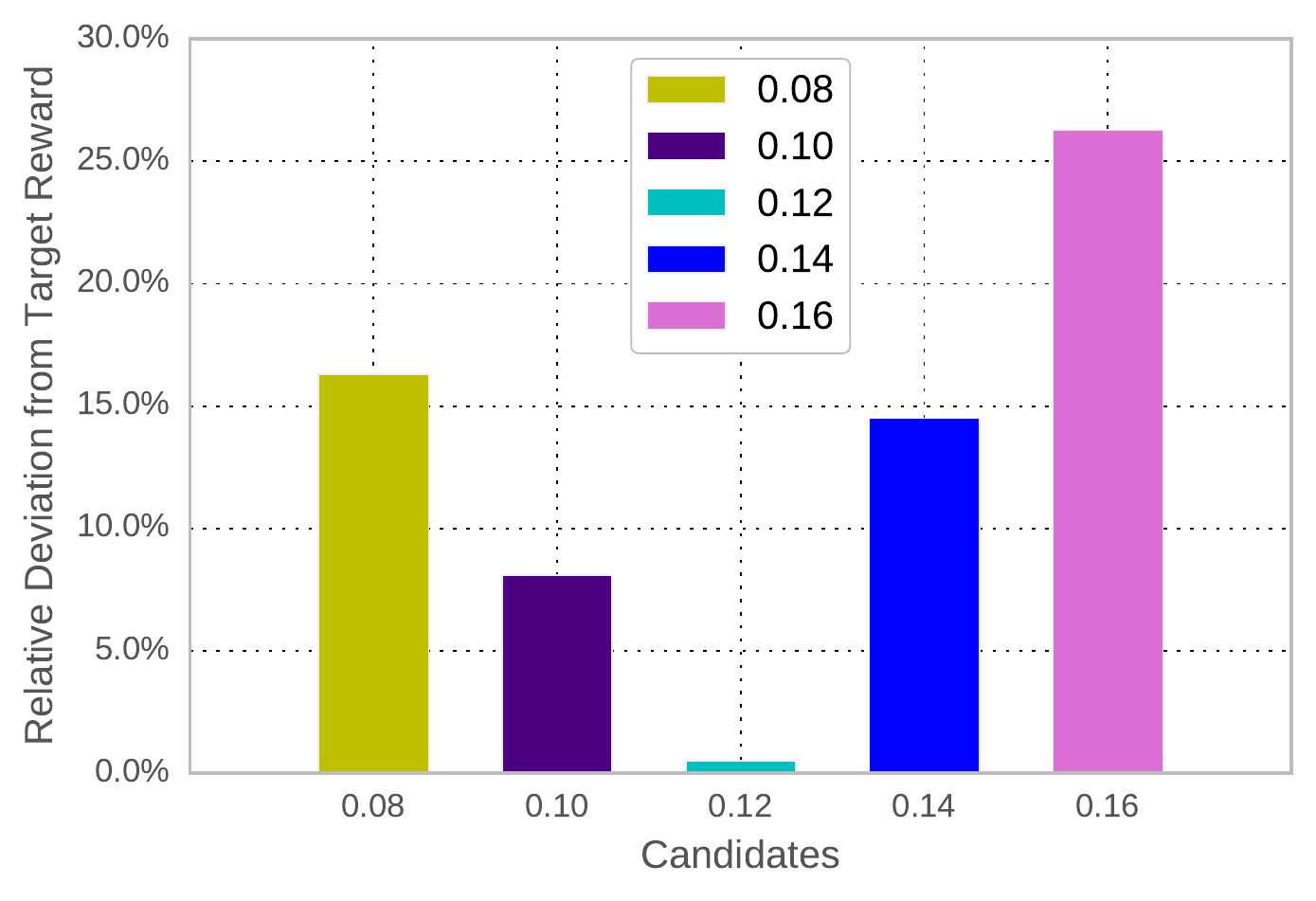}
         \caption{Relative deviation to the target reward for different candidates}
         \label{fig:ctr_deviation}
     \end{subfigure}
          \hfill
    \caption{Experiments with Explore-Exploit to online learn hyper-parameters for a machine learning model}
    \label{fig:exp}
\end{figure}

In this experiment, we use Explore-Exploit to automatically learn the threshold of a personalization model that can give us a target CTR of 0.11. Traditionally, this may take an engineer days of manual work. Especially when there are multiple models to tune, it could take weeks to finish. However, Explore-Exploit removes all the manual work. In Figure \ref{fig:ctr_display}, we show the percentage of displays for different threshold candidates after running Explore-Exploit for a few days. As we can see, at start, Explore-Exploit was exploring the candidates uniformly. Then after a few days, it converges to the candidate 0.12 without human intervention. To verify the correctness of the converged candidate, we also show the relative deviation to the target reward for different candidates in Figure \ref{fig:ctr_deviation}. As shown, the converged candidate 0.12 indeed gives the smallest deviation.

\section{Related Work}
The Explore-Exploit framework is built upon a significant amount of related works in multi-armed bandit algorithms, reinforcement learning and active learning. \cite{vermorel2005multi} and \cite{kuleshov2014algorithms} give a good overview and empirical analysis of different multi-armed bandit algorithms. These algorithms have been shown to be very useful for many real world problems such as recommendation \cite{li2010contextual, mcinerney2018explore} and ranking \cite{radlinski2008learning}. Within Explore-Exploit, we have implemented multiple multi-armed bandit algorithms including $\epsilon$-Greedy, UCB1, Thompson Sampling, etc. Active learning has also demonstrated great advantages in many use cases \cite{liu2017active, asghar2016deep}. In Explore-Exploit, we have built multiple active learning operators that work both in streaming setting \cite{chu2011unbiased, zhu2007active} and pool setting \cite{settles2008multiple}. 

Recently, automated machine learning (AutoML) systems \cite{mendoza2016towards, feurer2015efficient, zoph2016neural, falkner2018bohb} have received abundant attentions. Such systems focus on automatically finding good algorithms and hyper-parameters (e.g. neural architectures) given some predefined datasets. Explore-Exploit shares similar concepts but focuses on the online setting in a production environment.


\section{Acknowledgments}
We would like to thank Kavya Sajeev, Nanshu Wang, Nikunj Bajaj, Zoe Papakipos, Pooja Sethi, Yarik Markov, Rajen Subba and everyone else who have contributed to the design and implementation of Explore-Exploit.

\bibliographystyle{unsrt}
\bibliography{ref}

\end{document}